%% file: main.tex
\documentclass{article}
\usepackage[preprint]{corl_2024}
\usepackage{graphicx}
\usepackage{dblfloatfix}
\usepackage{url}
\usepackage{wrapfig}
\usepackage{booktabs}
\newif\iffigs
\usepackage{capt-of}  
\usepackage{cuted}
\usepackage[font=small,labelfont=bf]{caption}
\usepackage{subcaption}

\usepackage[fleqn]{amsmath}
\usepackage{float}
\usepackage{listings}
\usepackage{setspace}
\usepackage{graphicx}
\usepackage{mathrsfs}
\usepackage{amssymb}
\usepackage{nicefrac}
\usepackage{algorithm}
\usepackage{algorithmicx}
\usepackage[noend]{algpseudocode}

\usepackage{xcolor}
\usepackage{listings}
\usepackage{textcomp}
\definecolor{backcolour}{rgb}{0.95,0.95,0.92}
\lstdefinestyle{mystyle}{
    backgroundcolor=\color{backcolour},   
    basicstyle=\ttfamily\footnotesize,
    breakatwhitespace=true,         
    breaklines=true,                 
    captionpos=b,                    
    keepspaces=true, 
    keywords={},
    showstringspaces=false,
    showtabs=false,                  
    tabsize=2
}
\usepackage{pythonhighlight}
\usepackage{flushend}

\makeatletter
\newcommand\fs@spaceruled{\def\@fs@cfont{\bfseries}\let\@fs@capt\floatc@ruled
  \def\@fs@pre{\vspace{0.4\baselineskip}\hrule height.8pt depth0pt \kern2pt}%
  \def\@fs@post{\vspace{-0.4\baselineskip}\kern2pt\hrule\relax\vspace{-12pt}}%
  \def\@fs@mid{\kern2pt\hrule\kern2pt}%
  \let\@fs@iftopcapt\iftrue}
\makeatother

\usepackage{lipsum}
\usepackage{changepage}
\usepackage{cite}

\usepackage{graphicx}
\usepackage{etoolbox}

\title{\LARGE \bf Just Add Force for Contact-Rich Robot Policies} 
\author{William Xie, Stefan Caldararu, Nikolaus Correll\thanks{$^{1}$All authors are with the University of Colorado at Boulder, Boulder, CO. Corresponding email: {\tt\footnotesize wixi6454@colorado.edu}}}
\begin{document}
\maketitle
\begin{abstract}
Robot trajectories used for learning end-to-end robot policies typically contain end-effector and gripper position, workspace images, and language. Policies learned from such trajectories are unsuitable for delicate grasping, which require tightly coupled and precise gripper force and gripper position. We collect and make publically available 130 trajectories with force feedback of successful grasps on 30 unique objects. Our current-based method for sensing force, albeit noisy, is gripper-agnostic and requires no additional hardware. We train and evaluate two diffusion policies: one with (forceful) the collected force feedback and one without (position-only). We find that forceful policies are superior to position-only policies for delicate grasping and are able to generalize to unseen delicate objects, while reducing grasp policy latency by near 4x, relative to LLM-based methods. With our promising results on limited data, we hope to signal to others to consider investing in collecting force and other such tactile information in new datasets, enabling more robust, contact-rich manipulation in future robot foundation models. Our data, code, models, and videos are viewable at \color{blue}{\url{https://justaddforce.github.io/}}.
\end{abstract}

\vspace{-12pt}
\section{Introduction} \label{sec:intro}
Robot foundation models \citep{rt1, visualcortex, openx, octo, openvla, droid, ecot, tinyvla} leverage large-scale datasets spanning diverse objects, scenes, and embodiments to produce generalizable, cross-platform robot policies. The utilized data adheres to limited modalities: vision, language, and robot action--most typically, workspace camera view, text annotation of a given task, end-effector pose, and binary (open or closed) gripper position \citep{openx}. The latter, binary gripper position, especially without force feedback, precludes robot foundation models from successfully grasping many delicate objects such as soft produce, brittle dried goods, paper containers, and other such fragile and deformable items. In this paper, we propose a modification to this archetypal structure: continuous, rather than binary, gripper positions and corresponding grasp force feedback.

We contribute 1) a novel dataset of 130 trajectories with continuous gripper position and force feedback, spanning 30 unique objects ranging in deformability, volume, and mass (from 1g to 500g) and 2) train diffusion policies \citep{dp} with and without force feedback, showing that force enables delicate grasping performant with state-of-the-art LLM-based methods at a near 4x reduced latency with promise for generalizability at greater data scale.

Our position is that force, a strong supervisory signal of contact and grasp-success, along with continuous gripper position, rather than binary open or closed states, should be included in future datasets used in the training of robot foundation models. Our current-draw-based force sensing method is gripper-agnostic and requires no special hardware (``skin" or otherwise). While noisier and less accurate than bespoke solutions, policies trained on our data are capable of delicate grasps. Improved resolution and frequency of force and other tactile signals likely would further improve grasp fidelity and robustness.

\begin{figure*}[!th]
    \centering
    \includegraphics[width=1.0\textwidth]{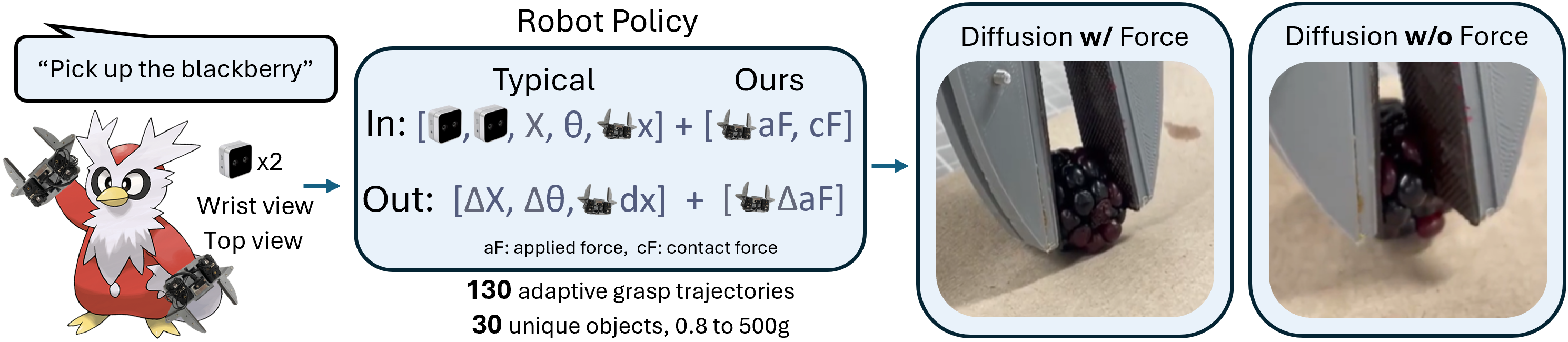}
    \captionof{figure}{We leverage LLM-directed expert demonstrations \citep{dg} of delicate objects to generate a dataset of 130 successful grasps of 30 different objects spanning a variety of physical properties. Our trajectories, unlike other datasets used in end-to-end learning \citep{openx, droid}, contain observed gripper applied and contact force and the action of increased gripper applied force. We train diffusion policies \citep{dp} on the dataset with and without force data and observe that forceful policies can, despite limited data, replicate trained behavior and generalize to unseen delicate objects at 4x reduced latency relative to LLM-policies, and position-only policies cannot. }
    \label{fig:overview}
    \vspace{-14pt}
\end{figure*}
\vspace{-6pt}
\section{Related Work} \label{sec:related}
\vspace{-6pt}
Large-scale robotic datasets \citep{openx, droid} have enabled the emergence of generalist, end-to-end robot foundation models \citep{rt1, visualcortex, octo, openvla, ecot, tinyvla} which typically append a behavior cloning architecture \citep{bet, act, vqbet, dp} to generate robot policies from a larger representation space. However, these robot foundation models are pre-trained on limited modalities: vision, language, and robot joint and/or end effector data.

There is a growing field exploring new modalities for end-to-end robot policy models, primarily in audio and tactile sensing \citep{maniwav, seehear, anyskin}. Such policies offer novel advantages in contact-rich manipulation and manipulation in visually occluded scenes but require new complexities, namely: custom and/or nontrivially emulated hardware and increased model complexity in processing and incorporation of high-dimension input data. In comparison, manipulator applied force and contact normal force can be approximated as 1-dimensional. Octo does explore finetuning on wrist force-torque for insertion tasks, but not grasp force \citep{octo}. And while traditional grasp force sensing is costly relative to audio and touch and thus unused in end-to-end learning, we leverage current draw as a gripper-agnostic force measurement, without additional sensing hardware, using a MAGPIE gripper \citep{dg, magpie} which interfaces with its motor control board to more easily provide this information.

In this work we examine grasping of delicate and deformable objects, which has primarily been done via adaptive grasping methods with traditional closed-loop control or LLM-based robot control: \citep{dg, adaptive_grasp, literally_us, switched_adaptive_grasp, tactile_adaptive_grasping}. Traditional controllers are not as generalizable as methods leveraging large amounts of data \citep{dg}, such as LLM-based methods, which in turn are high latency and computationally expensive. Utilizing force feedback from expert demonstrations of adaptive grasping in training or fine-tuning of robot foundation models may yield both lower latency and high generalizability.

\vspace{-10pt}
\section{Methods}
\vspace{-8  pt}
We introduce a dataset of 130 successful adaptive grasp trajectories across 30 unique objects spanning two orders of magnitude in mass (1g to 500g) and variable deformability (additional dataset detail and download link in \ref{appendix:peritem} and \ref{appendix:links}). Data is collected at 5 Hz from a MAGPIE gripper \citep{magpie} on a UR5 robot arm with a wrist-mounted Realsense D405 camera and a Realsense D435 camera overlooking a square, 55cm table. The user also provides a task instruction. The robot is positioned arbitrarily above and in-front of the target object, and the target object is placed arbitrarily on the table. We make our dataset publically available in an RLDS format \citep{rlds} compatible with Open-X and Droid datasets, Octo models, and other foundation models trained on RLDS format data.

To collect expert demonstrations, we employ DeliGrasp \citep{dg}, which navigates to the object and queries the LLM with the user-provided object description and uses LLM-estimated object mass, friction coefficient, and spring constants as parameters in a proportional controller which increases applied force and gripper closure until a measured contact force \citep{adaptive_grasp, literally_us}. We command applied force by incrementing motor torque limit on a Dynamixel motor (an equivalent actuator-agnostic approach would be to increase supply current), and we measure contact force from increased current draw.

We ``distill" these expert demonstrations \citep{scalingup} by training four diffusion policies \citep{droid, dp} on this data, with and without (position-only) force, and with the entire trajectory or the grasp-only (GO) (training details in \ref{appendix:training}). Initial testing showed that full trajectory policies did not learn meaningful robot motion, potentially due to the low amounts of data and each (robot start, target object) position pair being unique. Henceforth, we refer only to the policies trained on grasp-only data. By default, position-only policies apply a constant 2N and forceful policies begin at the lowest setting, 0.15N.

In our experiments we localize the object and position the robot at a viable grasp position using \citep{dg} and deploy and evaluate the policies only during the stationary grasp portion of a trajectory. We manually qualify deformation failures on a per-object common-sense basis (object crushed, cracked, etc...) and check for slip by raising the robot gripper directly vertically by 10cm. As the average adaptive grasp in the dataset completes in under 10 steps, for one ``grasp" we rollout the policy for 15 steps at 4Hz (3.75s per grasp vs 14.11s for an LLM-based grasp \citep{dg}, a 3.76x reduction).

\vspace{-6pt}
\section{Experiments} \label{sec:Results}
\vspace{-6pt}
We conduct 10 trials of grasps on 10 different objects: four objects seen in the training set (empty paper cup, raspberry, tomato, paper cup filled with water) but assessed to be difficult objects and six unseen objects (blackberry, egg, empty metal can, empty soft-shelled taco, pepper, potato chip). We compare between two models: 1) position-only policies (PO) with the canonical gripper position input and output and image \& task instruction inputs, and 2) forceful policies with applied force and contact force as additional inputs and applied force as an additional output.
\begin{figure*}[!th]
    \centering
    \includegraphics[width=1.0\textwidth]{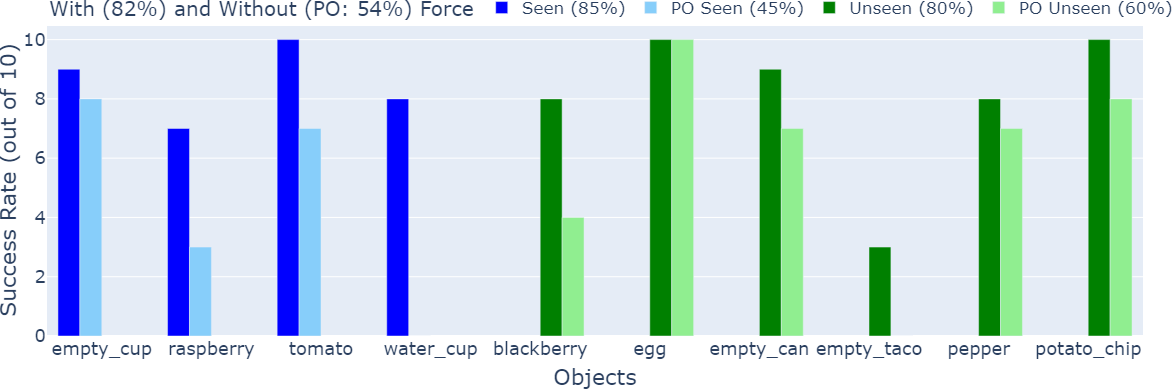}
    \captionof{figure}{We conduct a series of 10 trials for a selection of 10 objects; four seen in training, six unseen. Forceful policies (82\%) replicate seen grasps (85\%) and generalize to similar but unseen objects (80\%). position-only policies (54\%) retain a level of performance on seen (45\%) and improve on unseen (60\%) delicate objects, suggesting that continuous gripper position control alone contributes to successful delicate grasps. We note that position-only policy failures are generally deforming and compress more than forceful policies (see Fig. \ref{fig:breakdown})}
    \label{fig:barchart}
    \vspace{-12pt}
\end{figure*}

\begin{figure*}[!th]
    \centering
    \includegraphics[width=1.04\textwidth]{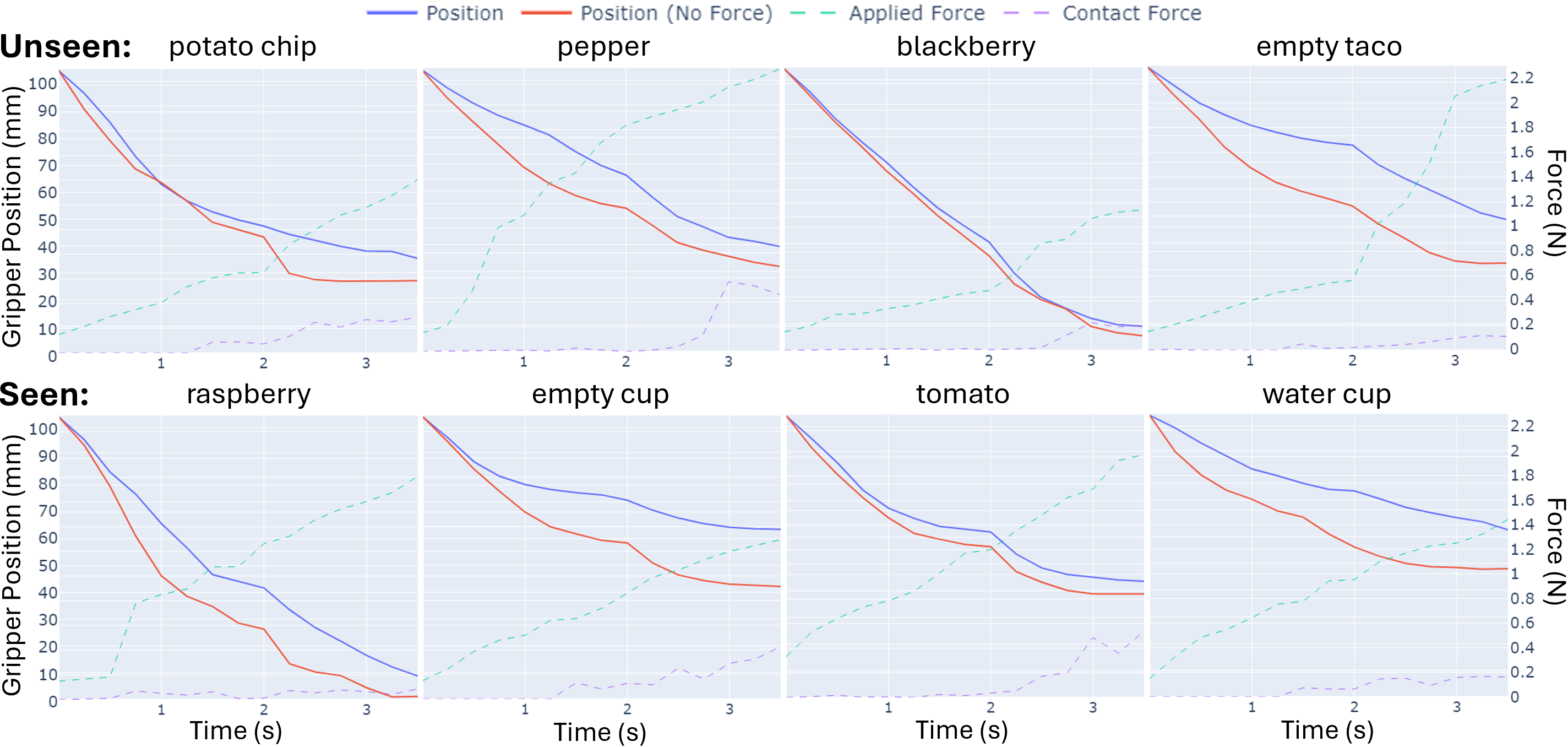}
    \captionof{figure}{We plot 1) forceful policies gripper position (blue), applied force (green dash), and contact force (purple dash) and 2) position-only policies gripper position (red) against time, with additional plots in \ref{appendix:extra_unseen}. Uniformly, position-only policies close more narrowly than forceful policies, leading to deformation failures, particularly for delicate objects like blackberries and raspberries. Individual position-only policy grasps on produce like tomatoes and peppers are successful, but we observe that after 10 trials, the produce is noticeably deformed due to greater compression, unlike for forceful grasps. On objects like the pepper, empty taco, blackberry, and tomato, applied force flattens as contact force increases.}
    \label{fig:breakdown}
    \vspace{-6pt}
\end{figure*}

Across all objects, we find that forceful policies (82\% success) are superior to position-only policies (54\% success) (Fig. \ref{fig:barchart}) and that position-only policies compress more than forceful policies (Fig. \ref{fig:breakdown}). Position-only policies are still capable, perhaps because they are artifacts of forceful adaptive grasping, just trained without the force feedback, and the control law may be implicitly learned through solely vision, gripper position, and task instruction. Forceful policies generalize to unseen objects (80\% success, compared to 85\% for seen objects) and withhheld policies improve (60\%, up from 45\%), potentially due to relatively stiff objects like the egg and potato chip being forgiving for additional compression.

More granularly, we qualify failures as either deformation or slip. While both policies generally perform deformation failures, forceful policies slip (7 occurrences) more than position-only policies (3 occurrences), representing a 28\% vs. 6.1\% share of respective policy failures. For produce like tomatoes and peppers, position-only policies generate grasps which are individually successful, but we observe that after 10 trials, the produce is noticeably deformed (``mushy") due to repeated greater compression, unlike for forceful policies (Fig. \ref{fig:breakdown}). We leave these grasps marked as successes as the produce ``mushy" threshold of desirability is dependent on the end-user. 

Additionally, both policies occasionally generate generated grasps which terminate several mm, up to several cm, offset from the object. We note these occurrences as ``null grasps," separate from successes or failures. We note that the forceful policies produced null grasps 11.5\% of the time (13 occurrences, even across seen and unseen grasps) and position-only policies produced null grasps 20\% of the time (25 occurrences with 6 on the raspberry and 5 on the blackberry). We also observe volatility, though much rarer, in rapidly increasing gripper position and force post-contact, resulting in abrupt crushes (notably affecting the average applied force on the raspberry in Fig. \ref{fig:breakdown}).

In Fig. \ref{fig:breakdown}, we depict per-object grasp trajectories and forces and observe that position-only policies uniformly compress more than forceful policies. Position-only policies are initially more aggressive in closing the gripper and often continue aggressive closure past contact, resulting in deformation failures. Forceful policies flatten applied force as contact force increases for some objects (pepper, empty taco, blackberry, tomato), showing vestiges of the proportional control law used in expert demonstrations, however, policies still apply more force than is typically needed and have not fully learned the control characteristics. Additionally, while objects span a large range of gripper position (5 to 65mm), final applied force lies in a smaller range (1.1N to 2.3N).
\vspace{-6pt}
\section{Conclusion} \label{sec:conclusion}
\vspace{-6pt}
We add force observations and actions to the common data structure of imagery, task instruction, robot pose, gripper position used in training end-to-end robot policy models in a dataset of 130 grasps across 30 objects. We train a diffusion policy trained on force feedback which outperforms a policy trained without force on delicate objects and generalizes to unseen objects, indicating that force may be a worthwhile inclusion in future data collection endeavors.

\textbf{Limitations and Future Work:} As the second derivative of gripper position, force may encode enough information to be all you need for manipulation. Our models are currently only evaluated at rest, and we do not explore adaptive grasping while in motion.  Moreover, our evaluated models are simplistic and trained on a toy dataset---future work includes finetuning on foundation models which allow new modalities \citep{octo} or collecting diverse, large scale data with force feedback. Adaptive grasping may also benefit from a pretrained LLM backbone to leverage common-sense reasoning about forces. Force also has applications beyond our demonstrated use case of slip/contact sensing and may be used for generating non-prehensile manipulation trajectories.

\clearpage
\bibliography{dg}

\clearpage
\onecolumn
\appendix
\section{Appendix} \label{sec:appendix}
\input{tex/appendix}

\end{document}

%% file: tex/appendix.tex
\subsection{Dataset Details}\label{appendix:peritem}
Objects: orange bottle, peeled garlic clove, stuffed animal, garlic clove, green block, tomato, red screwdriver handle, scallion stalk, small avocado, yellow ducky, water bottle, small black motor, empty paper cup, circuit board, red button, scalion stalk, orange noodle bag, yellow block, strawberry, bottle cap, small suction cup, light green chip, ziptie bag, metal lock, cardboard box, raspberry, large bearing, paper cup with water, small red green apple, paper airplane, green circuit board, plastic bottle, cherry tomato, mushroom, garlic bulb.

For most objects collect 5-7 trajectories, with a few one-offs. We use a webapp console to interoperate between DeliGrasp, robot control, and diffusion policy evaluation.
\begin{figure*}[!th]
    \centering
    \includegraphics[width=1.0\textwidth]{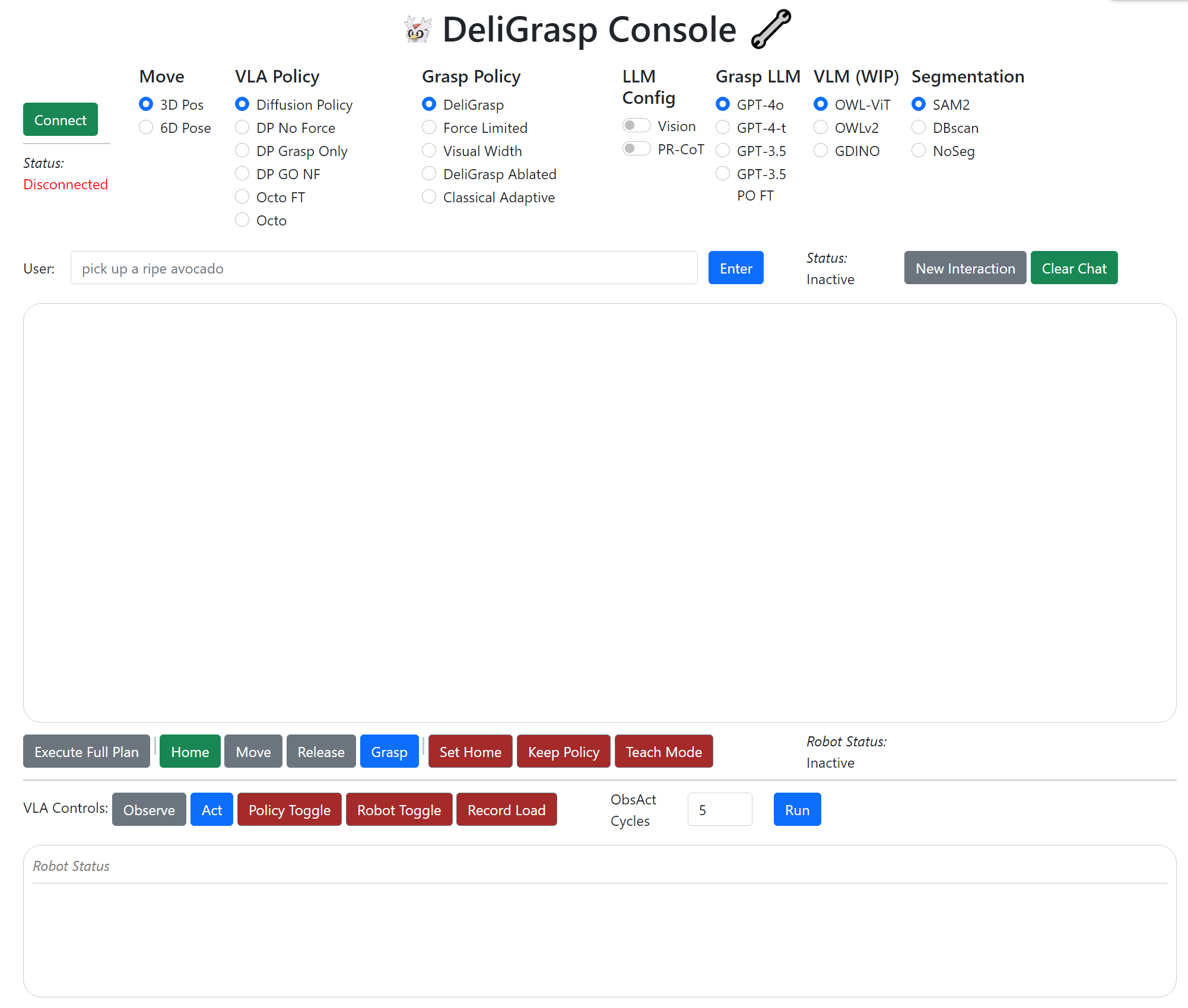}
\end{figure*}

Data is in the following TFDS format:
\begin{lstlisting}
FeaturesDict({
    'episode_metadata': FeaturesDict({
        'file_path': Text(shape=(), dtype=string),
    }),
    'steps': Dataset({
        'action': Tensor(shape=(9,), dtype=float64),
        'action_dict': FeaturesDict({
            'cartesian_position': Tensor(shape=(6,), dtype=float64),
            'gripper_force': Tensor(shape=(1,), dtype=float64),
            'gripper_position': Tensor(shape=(1,), dtype=float64),
            'rotation': Tensor(shape=(3,), dtype=float64),
            'translation': Tensor(shape=(3,), dtype=float64),
        }),
        'discount': Scalar(shape=(), dtype=float32),
        'is_first': Scalar(shape=(), dtype=bool),
        'is_last': Scalar(shape=(), dtype=bool),
        'is_terminal': Scalar(shape=(), dtype=bool),
        'language_embedding': Tensor(shape=(512,), dtype=float32),
        'language_instruction': Text(shape=(), dtype=string),
        'observation': FeaturesDict({
            'state': Tensor(shape=(16,), dtype=float64),
            'applied_force': Tensor(shape=(1,), dtype=float64),
            'cartesian_position': Tensor(shape=(6,), dtype=float64),
            'contact_force': Tensor(shape=(1,), dtype=float64),
            'gripper_position': Tensor(shape=(1,), dtype=float64),
            'image': Image(shape=(480, 640, 3), dtype=uint8),
            'joint_position': Tensor(shape=(6,), dtype=float64),
            'wrist_image': Image(shape=(480, 640, 3), dtype=uint8),
        }),
        'reward': Scalar(shape=(), dtype=float32),
        'subtask': Text(shape=(), dtype=string),
    }),
})
\end{lstlisting}
We the ``action\_dict" and observation keys after ``state" for compatibility with DROID. The ``subtask" key denotes, within a trajectory, whether the robot is moving toward, grasping, or returning home from an object. The grasp-only dataset is the grasping subset of trajectories.

\subsection{Diffusion Policy Training}\label{appendix:training}
We train our models using DROID Policy Learning \citep{droid}, which deviates from the vanilla implementations in three ways: 1) opting out of SparseSoftmax to retrieve regional keypoints, instead keeping the feature channels of the image embedding, 2) adding language conditioning by encoding task instruction and adding it to the observation input, and 3) downsizing the input dimension to a fixed size. We do not alter the DROID hyperparameters except for the following: we train for 3000 steps (30 epochs) and a batch size of 16. Model training is done locally on a 2070 Super, taking approximately 1 hour to train per model. We use $T_o$, $T_a$, $T_p$ of 2, 8, and 16, but in evaluation use receding horizon control ($T_a$ = 1).

\subsection{Data and Model Downloads}\label{appendix:links}
\begin{enumerate}
    \item \color{blue}{\url{https://justaddforce.github.io/datasets}}
    \item \color{blue}{\url{https://justaddforce.github.io/models}}
\end{enumerate}

\subsection{Additional Unseen Plots}\label{appendix:extra_unseen}
\begin{figure*}[!th]
    \centering
    \includegraphics[width=0.7\textwidth]{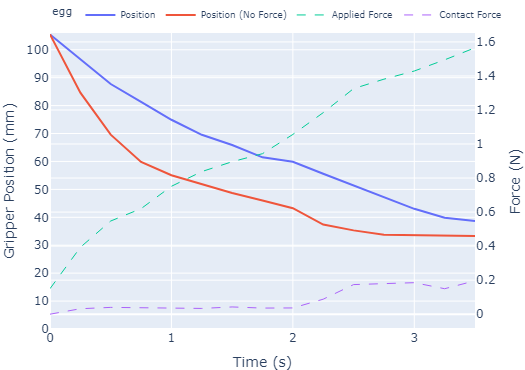}
\end{figure*}

\begin{figure*}[!th]
    \centering
    \includegraphics[width=0.7\textwidth]{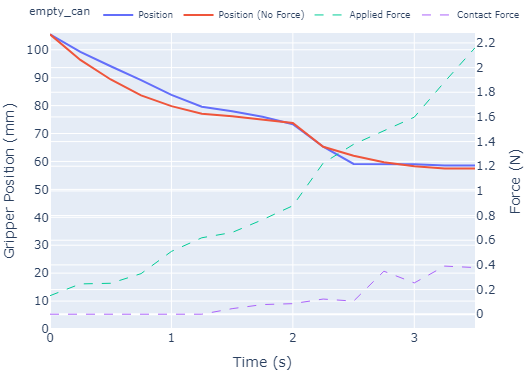}
\end{figure*}